\documentclass[conference]{IEEEtran}

\usepackage{cite}
\usepackage{amsmath,amssymb,amsfonts}
\usepackage{algorithmic}
\usepackage{graphicx}
\usepackage{textcomp}
\usepackage{xcolor}
\usepackage{booktabs}
\usepackage{multirow}
\usepackage{url}
\usepackage{subcaption}
\usepackage{hyperref}
\usepackage{float} 

\begin{document}

\title{D3R-Net: Dual-Domain Denoising Reconstruction Network for Robust Industrial Anomaly Detection}

\author{
    \IEEEauthorblockN{Dmytro Filatov}
    \IEEEauthorblockA{\textit{Applied Artificial Intelligence and Computer Vision} \\
    \textit{Aimech Technologies Corp.}\\
    San Francisco, The United States of America \\
    dima@deepxhub.com}
    \and
    \IEEEauthorblockN{Valentyn Fedorov}
    \IEEEauthorblockA{\textit{Artificial Intelligence and Computer Vision} \\
    \textit{Aimech Technologies Corp.}\\
    Kyiv, Ukraine \\
    valentyn.fedorov@deepxhub.com}
    \and
    \IEEEauthorblockN{Vira Filatova}
    \IEEEauthorblockA{\textit{Applied Artificial Intelligence} \\
    \textit{Covijn Ltd.}\\
    London, United Kingdom \\
    vira@deepxhub.com}
    \and
    \IEEEauthorblockN{Andrii Zelenchuk}
    \IEEEauthorblockA{\textit{Artificial Intelligence and Computer Vision} \\
    \textit{Aimech Technologies Corp.}\\
    Kyiv, Ukraine \\
    andrii.zelenchuk@deepxhub.com}
}

\maketitle

\begin{abstract}
Unsupervised anomaly detection (UAD) is a key ingredient of automated visual inspection in modern manufacturing. The reconstruction-based methods appeal because they have basic architectural design and they process data quickly but they produce oversmoothed results for high-frequency details. As a result, subtle defects are partially reconstructed rather than highlighted, which limits segmentation accuracy. We build on this line of work and introduce D3R-Net, a Dual-Domain Denoising Reconstruction framework that couples a self-supervised ``healing'' task with frequency-aware regularization. During training, the network receives synthetically corrupted normal images and is asked to reconstruct the clean targets, which prevents trivial identity mapping and pushes the model to learn the manifold of defect-free textures. In addition to the spatial mean squared error, we employ a Fast Fourier Transform (FFT) magnitude loss that encourages consistency in the frequency domain. The implementation also allows an optional structural similarity (SSIM) term, which we study in an ablation. On the MVTec AD Hazelnut benchmark, D3R-Net with the FFT loss improves localization consistency over a spatial-only baseline: PRO AUC increases from 0.603 to 0.687, while image-level ROC AUC remains robust. Evaluated across fifteen MVTec categories, the FFT variant raises the average pixel ROC AUC from 0.733 to 0.751 and PRO AUC from 0.417 to 0.468 compared to the MSE-only baseline, at roughly 20 FPS on a single GPU. The network is trained from scratch and uses a lightweight convolutional autoencoder backbone, providing a practical alternative to heavy pre-trained feature embedding methods.
\end{abstract}


\section{Introduction}
\label{sec:intro}

\IEEEPARstart{V}{isual} quality inspection is a standard component of automated manufacturing lines. In many industrial scenarios, defective parts are rare, diverse, and costly to annotate, which makes fully supervised defect segmentation hard to scale. A more practical setup is unsupervised anomaly detection (UAD), where models are trained only on defect-free samples and must flag deviations from this normal data distribution at test time. Reconstruction-based methods, typically built around autoencoders or U-Net-style decoders, assume that a model fitted on normal data will reproduce normal regions well but fail to reconstruct unseen defects. The absolute reconstruction error is then used as an anomaly map. Despite their simplicity and favorable runtime, plain convolutional autoencoders trained with pixel-wise losses (e.g., $L_2$) have a well-known tendency to blur high-frequency content. When the model learns to predict an average of plausible textures, small scratches, cracks, or holes can be partially ``washed out'' in the reconstruction and thus less pronounced in the residual. Many industrial defects, however, manifest primarily as local perturbations of texture statistics and edges, which are easier to interpret in the frequency domain than in RGB space alone.

At the same time, simply increasing network capacity or switching to more complex backbones moves reconstruction-style approaches closer to feature embedding methods that rely on large ImageNet-pretrained networks, losing some of their practical advantages. In this work we revisit convolutional autoencoder baselines and strengthen them in two ways. First, we adopt a self-supervised denoising (``healing'') task: the model receives images where localized synthetic defects are injected and must reconstruct the original clean images. This construction discourages a shortcut solution that copies the input and encourages the network to learn the structure of defect-free textures. Second, we enforce dual-domain consistency by penalizing discrepancies between the magnitude spectra of the ground truth and reconstruction, complementing the usual spatial reconstruction loss. Beyond a single-class case, we embed D3R-Net into a unified benchmarking pipeline~\cite{pleskach_methods} that evaluates multiple methods across the full MVTec AD dataset. This pipeline computes both image-level and pixel-level metrics as well as PRO AUC and inference throughput, allowing us to assess not only segmentation quality but also runtime characteristics.

\noindent \textbf{Main Contributions.}
\begin{enumerate}
    \item We formulate D3R-Net, a dual-domain denoising reconstruction model trained on a self-supervised healing task with localized synthetic defects, including blended foreign patches that mimic realistic texture changes.
    \item We combine the spatial mean squared error with an FFT-based magnitude loss, and optionally a structural similarity term, to better preserve fine-grained textures without increasing backbone complexity.
    \item We integrate D3R-Net into a common evaluation pipeline~\cite{pleskach_methods} together with autoencoder, PaDiM, STFPM and PatchCore baselines, and report both per-class and average results on the MVTec AD dataset. On Hazelnut, the FFT-regularized variant yields a substantial absolute gain in PRO AUC over the MSE-only D3R baseline.
\end{enumerate}

The remainder of this paper is organized as follows. Section~\ref{sec:related} reviews related work in reconstruction-based and feature-embedding anomaly detection. Section~\ref{sec:method} details the proposed methodology, including the synthetic corruption strategy and the dual-domain loss functions. Section~\ref{sec:experiments} describes the experimental setup, dataset, and evaluation metrics. Section~\ref{sec:results} presents the quantitative and qualitative results on the MVTec AD dataset. Finally, Section~\ref{sec:concl} concludes the paper and discusses future research directions.

\section{Related Work}
\label{sec:related}

In this section we briefly review three main families of methods that are relevant for our work and clarify where D3R-Net fits among them.

\subsection{Reconstruction-Based Methods and the Blurriness Issue}
The most direct way to approach unsupervised anomaly detection is to learn a model that simply reconstructs normal images and uses the reconstruction error as an anomaly score. Early works in this line used convolutional autoencoders trained with a pixel-wise mean squared error loss. The intuition is straightforward: if the model has only seen defect-free samples during training, it should struggle to reproduce unseen defects at test time. In practice the picture is more complicated. When a network is optimized purely for an $L_2$ type reconstruction loss, it tends to favour the conditional mean of the data distribution given its latent representation. If several fine-scale textures are compatible with the same coarse structure, the mean of these textures is typically smoother than any of them. As a result, thin scratches, hairline cracks or small holes are often partially ``filled in'' during reconstruction. The corresponding residual maps become less peaked and harder to threshold. A number of works tried to compensate for this effect. Structural-similarity-based losses have been added on top of MSE to put more emphasis on local contrast and perceived structure~\cite{ssim_ae}. Adversarial training has also been explored, for example in GANomaly~\cite{ganomaly}, where a generative adversarial network encourages sharper outputs and a discriminator helps shape the latent space. These approaches can improve visual quality but often at the price of more fragile training and a higher implementation burden. Our work stays on the reconstruction side but follows a different route. The model preserves its fundamental design through fundamental adversarial training which enables it to measure spectral differences between reconstructed data and actual data through its dual-domain loss function. The training objective enables us to handle blurriness through denoising-style training while maintaining the existing architecture structure.

\subsection{Feature Embedding Approaches}
The current state of the art on MVTec AD is dominated by feature-embedding methods such as PaDiM~\cite{padim}, PatchCore~\cite{patchcore} and SimpleNet~\cite{simplenet}. The methods discard traditional reconstruction approaches because they use ImageNet pre-trained deep backbone features from ResNet or WideResNet variants to extract information. PaDiM models the distribution of features at each spatial location as a Gaussian and measures Mahalanobis distance at test time. PatchCore builds a memory bank of patch-level features and uses nearest-neighbour distances in that space as anomaly scores. STFPM~\cite{stfpm} uses a student-teacher setup where the student network is trained to match a frozen teacher's multi-scale features on normal data, and deviations in the feature maps are interpreted as anomalies. These methods are very effective and achieve impressive numbers on most benchmarks, but they also come with their own set of trade-offs. First, they rely on heavy backbones with tens of millions of parameters, which may be overkill in applications with tight compute or memory budgets. Second, they store a large number of feature vectors or covariance matrices, leading to non-negligible memory footprints. Third, their performance hinges on the assumption that ImageNet pre-training provides a useful inductive bias for industrial textures and objects, which may not always hold when the target domain departs significantly from natural images. D3R-Net deliberately sits at the other end of this spectrum. It uses a lightweight convolutional autoencoder (around 1-2M parameters), is trained from scratch on the target data and does not depend on any external datasets. The goal is not to beat all feature-embedding methods on every metric, but to provide a strong reconstruction-based baseline that remains attractive when pre-trained backbones are not an option.

\subsection{Synthetic Anomaly Training and Frequency-Domain Techniques}
Another strand of work explores the idea of training on synthetic anomalies. CutPaste~\cite{cutpaste} showed that simply cutting and pasting image patches can provide a useful self-supervised signal: a classifier trained to distinguish real from patched images learns features that transfer well to anomaly detection. DRAEM~\cite{draem} went further and introduced a reconstruction-based framework where one network learns to inpaint synthetically corrupted regions, while a separate discriminative head is trained to segment those regions. These approaches highlight two important points. First, synthetic defects need not be perfectly realistic to be useful; it is enough that they encourage the model to learn the statistics of clean normal regions. Second, explicitly learning to undo local corruptions can act as a regularizer that prevents trivial identity mappings. Our setting is close in spirit but intentionally simpler. We use on-the-fly synthetic corruptions as a self-supervised ``healing'' task, but we do not introduce an additional discriminative branch. The same autoencoder is trained to remove corruptions and reconstruct clean images, and the anomaly score is derived directly from the reconstruction residual. On top of that, we add a frequency-domain consistency term: while Fourier analysis has been used in other vision tasks and explored in recent anomaly detection work, we focus on a particularly lightweight formulation that fits naturally into the training loop of a standard autoencoder and can be implemented with a few lines of code.

\section{Methodology}
\label{sec:method}

\begin{figure*}[t!]
    \centering
    \includegraphics[width=0.85\textwidth]{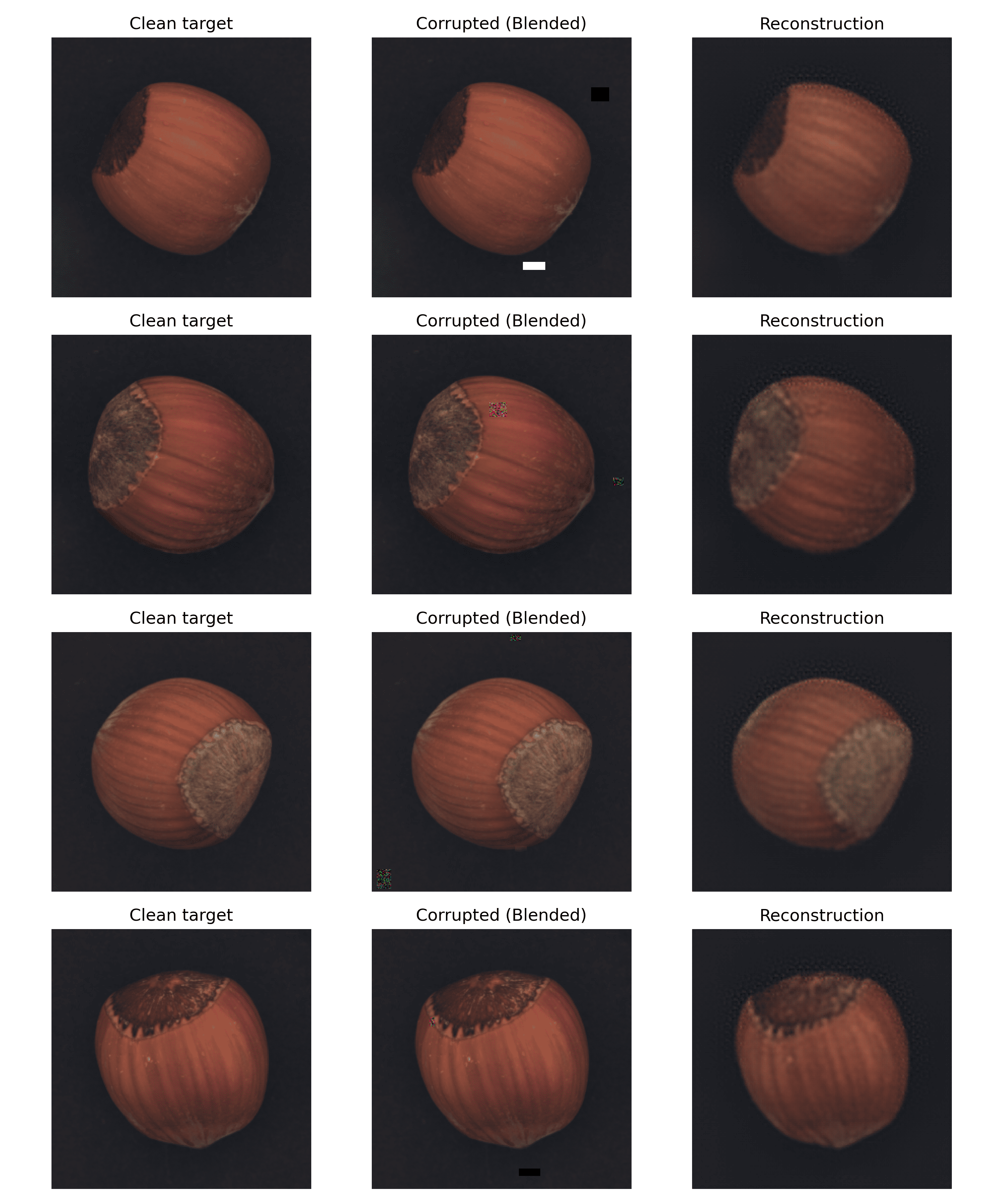}
    \caption{\textbf{Self-supervised healing task for D3R-Net (multi-category benchmark run).} Each row shows a clean hazelnut image, its synthetically corrupted version, and the reconstruction produced by D3R-FFT. Corruptions include local occlusions, noise patches and blended foreign patches, and the model is trained to remove them and recover the normal texture.}
    \label{fig:healing_demo}
\end{figure*}

\subsection{Overview}
D3R-Net is trained exclusively on normal images. Given a clean sample $x$, we apply a corruption operator $\mathcal{C}$ to obtain a degraded input. A convolutional autoencoder is then optimized to reconstruct the original target. Because the network never observes real defects during training, its capacity is devoted to modelling the manifold of defect-free textures. At test time we feed uncorrupted images through the same network and measure the discrepancy between input and reconstruction as an anomaly score. A standard argument from denoising theory is useful here. If we restrict attention to the mean squared error part of the loss and assume that the network is expressive enough, then the optimal reconstruction for a fixed corruption process satisfies the conditional expectation of a clean image given its corrupted version. Intuitively, the network learns what a ``typical'' clean patch should look like under the local context of the corrupted input. When we apply the same model to truly anomalous regions at test time, those regions no longer resemble anything from the normal training distribution, so the conditional expectation and the input disagree and the residual becomes large.

\subsection{On-the-Fly Synthetic Corruption}
We do not pre-generate a separate dataset of corrupted images. Instead, corruptions are applied on-the-fly inside the training loop. The system provides two operational benefits because it uses minimal memory resources and shows different corrupted versions of the same image during each training epoch which reduces the risk of artifact-specific model overfitting. If corruption is applied, we sample an integer $K$ and generate $K$ localized defect regions. For each region we sample a rectangle with random position and size. The side lengths are drawn from a small range so that each patch occupies only a few percent of the image area. For each region we then choose one of three corruption types: 
\\
a) \textbf{Rectangular Occlusion}: We overwrite the pixels in the region with a constant intensity. This simulates missing material, specular highlights or saturated sensor regions. 
\\
b) \textbf{Gaussian Noise Patch}: We draw a tensor of Gaussian noise with zero mean and standard deviation on the region and add it to the image, with clipping. This produces local grainy patterns that disturb texture statistics without changing the global layout. 
\\
c) \textbf{Foreign Patch Interpolation}: We randomly choose another image from the same minibatch and extract its content on the region. Instead of pasting it directly, we perform a convex combination with a random mixing coefficient. This simple ``foreign patch interpolation'' produces defects that look more like stains, discolorations or subtle texture changes than like hard-edged blocks. The resulting patterns are still clearly off-manifold from the point of view of the normal data, but they avoid the trivial case where the network could solve the task by detecting sharp cut boundaries. It is worth stressing that we do not require these synthetic defects to be indistinguishable from real ones. The autoencoder needs to learn typical texture patterns instead of copying every possible anomaly for this objective.

\subsection{Network Architecture}
The reconstruction backbone is a compact convolutional autoencoder. We denote the encoder by $E$ and the decoder by $D$. In the implementation we use four downsampling blocks in the encoder and four corresponding upsampling blocks in the decoder. The encoder block contains a $4 \times 4$ Conv2D layer which performs stride 2 operations with padding 1 followed by Batch Normalization and ReLU activation. The model produces a $16 \times 16$ latent map after it processes $256 \times 256$ input data. Each decoder block mirrors the encoder using ConvTranspose2D with the same kernel and stride, again followed by Batch Normalization and ReLU, except for the final layer which ends with a sigmoid to produce outputs in $[0, 1]$. The number of channels increases with depth in the encoder and decreases symmetrically in the decoder. The model maintains its basic structure because we chose to remove skip connections and multi-branch designs which allows for quick model training and compact model dimensions.

\subsection{Dual-Domain Loss and Frequency Normalization}

\textbf{Spatial Reconstruction Term}: The first component of the loss is the standard mean squared error between the input and the reconstruction. This term enforces global structural correctness and colour consistency. On its own, however, it does not prevent the model from smoothing out ambiguous high-frequency details.

\textbf{Frequency-Domain Term and Dynamic Range}: To complement the spatial term we introduce a loss in the frequency domain. Let $\mathcal{F}$ denote the 2D Fast Fourier Transform applied channel-wise. For each channel we compute its magnitude spectra. The FFT loss is defined as the $L_1$ norm between the magnitudes. One technical detail is how to handle the very different scales of low and high frequencies. It is common to use a logarithmic transform when visualizing spectra, because most of the energy sits in the low-frequency coefficients. We experimented with a similar idea for the loss, but found that in our setting a direct $L_1$ penalty on the raw magnitudes leads to more stable improvements in localization. Intuitively, the log transform compresses differences in the high-frequency tail too aggressively, which makes the loss less sensitive to exactly the structures we care about (small cracks, sharp edges, textured regions). Using the unscaled magnitude keeps these differences visible to the optimizer. In the implementation we use a normalized FFT (orthonormal transform), which keeps the overall scale of the spectra manageable and avoids manual rescaling of gradients.

\textbf{Optional Structural Similarity Term}: The codebase also supports an SSIM-based term. This term is sensitive to local luminance, contrast and structure and can be seen as sitting between the purely spatial MSE and the global spectral loss. In most of our experiments we set its weight to zero and treat the SSIM-augmented variant as an ablation.

\textbf{Total Objective}: We minimize the expectation of the total loss over the joint distribution of clean and corrupted images. Unless otherwise stated, we use specific weights for FFT and SSIM terms.

\subsection{Anomaly Scoring and Visualization}
At test time we no longer corrupt the input. For each test image we compute its reconstruction and define the per-pixel anomaly map as the difference between the input and the reconstruction. These maps are used directly to compute pixel-level ROC and PR curves and to derive PRO AUC. For image-level detection we reduce the map to a scalar score via a simple reduction such as the maximum over spatial locations.

\section{Experiments}
\label{sec:experiments}

\subsection{Dataset and Task}
All experiments are conducted on the MVTec AD dataset~\cite{mvtec}. It comprises 15 industrial object and texture categories: bottle, cable, capsule, carpet, grid, hazelnut, leather, metal\_nut, pill, screw, tile, toothbrush, transistor, wood, zipper. For each category, the training split contains only defect-free images, while the test split includes both normal and defective samples together with pixel-level annotations for all anomalous regions. We adhere to the official train/test splits without using an additional validation subset. Models are trained and evaluated independently for each category. This setting reflects a realistic deployment scenario, where one typically tunes and maintains a separate detector per product type or surface. Conceptually, some categories are dominated by clearly delineated objects (e.g., hazelnut, bottle, capsule), where the model must reconstruct a relatively compact foreground object against a simpler background. Others are essentially textures (leather, tile, wood, carpet), where the main challenge is to reproduce a stochastic but stationary pattern over the whole image. We will refer to both types when discussing the results.

\subsection{Compared Methods}
We evaluate the proposed D3R-Net variants against both reconstruction-based and feature-embedding baselines within the unified pipeline described in~\cite{pleskach_methods}:
\\
\textbf{AE-MSE}: a plain convolutional autoencoder trained on uncorrupted normal images with a mean squared error loss. This serves as a reference reconstruction baseline.
\\
\textbf{D3R-MSE}: the same autoencoder architecture trained on the self-supervised healing task with synthetic corruptions, using only the MSE term in the loss. This isolates the effect of the denoising formulation.
\\
\textbf{D3R-FFT}: the full model with healing task and dual-domain loss, combining MSE and FFT magnitude discrepancy.
\\
\textbf{D3R-FFT-SSIM}: an extended configuration that additionally includes an SSIM-based term. Its role is mainly ablation; it is not the primary recommended setting.
\\
\textbf{PaDiM}~\cite{padim}: a patch-based density model on ImageNet-pretrained ResNet-18 features. It models per-location Gaussian distributions in feature space.
\\
\textbf{STFPM}~\cite{stfpm}: a student-teacher feature pyramid matching approach, where the teacher is a frozen ResNet-18 and the student learns to mimic its multi-scale activation patterns on normal data.
\\
\textbf{PatchCoreLite}~\cite{patchcore}: a memory-reduced variant of PatchCore with a ResNet-18 backbone. It approximates the full PatchCore behaviour while remaining more tractable in memory and runtime.
\\
The reconstruction-based methods are trained from scratch on each category. PaDiM, STFPM and PatchCoreLite reuse ImageNet-pretrained backbones and thus benefit from external semantic information.

\subsection{Training and Implementation Details}
For AE-MSE, D3R-MSE, D3R-FFT and D3R-FFT-SSIM we use the same convolutional autoencoder described in Section III. All reconstruction-based models share the following training setup unless noted otherwise: Optimizer: Adam with learning rate $10^{-3}$. Batch size: 8. Epochs: 50 per category. Input size: images resized to $256 \times 256$ and normalized to $[0, 1]$. Corruption probability: 0.5 for D3R variants, with up to three corrupted regions per image and patch sizes sampled in a fixed range. For the D3R variants, corruptions are applied on-the-fly at each iteration. The corruption parameters are kept constant across all categories. PaDiM, STFPM and PatchCoreLite follow their respective reference implementations. For STFPM, the student network is trained for 30 epochs with a learning rate of $10^{-4}$. For PaDiM and PatchCoreLite, feature extraction is done on the training set only, and the resulting statistics or memory banks are used at test time without further fine-tuning. All methods are trained and evaluated with the same random seed for fair comparison. We do not perform any manual per-category hyperparameter tuning; the intention is to assess how each method behaves under a single, fixed recipe across all 15 categories.

\subsection{Evaluation Metrics}
We evaluate models at both image and pixel levels and also measure inference throughput. 
\\
a) \textbf{Image-Level Metrics}: Each method produces a scalar anomaly score per test image. We treat this as a binary classifier between normal and defective images and compute: ROC AUC (the area under the receiver operating characteristic curve) and AP (Average Precision). 
\\
b) \textbf{Pixel-Level Metrics}: From the anomaly map we obtain a score per pixel. Flattening the map and the corresponding ground truth masks over all test images yields vectors of scores and binary labels, from which we compute: Pixel ROC AUC and Pixel AP. These metrics capture how well the method distinguishes anomalous pixels from normal ones across the entire dataset. 
\\
c) \textbf{Per-Region Overlap (PRO) AUC}: To better reflect the quality of spatial localization, we adopt the PRO AUC metric used in~\cite{mvtec}. Given a threshold on the normalized anomaly maps, we obtain a binary prediction mask and measure: the false positive rate (FPR) as the fraction of normal pixels incorrectly marked as anomalous; and for each connected ground truth component, the overlap between the component and the predicted anomalous region, averaged over all components. Sweeping over thresholds produces a PRO-FPR curve. Following~\cite{mvtec}, we integrate PRO with respect to FPR up to a maximum FPR of 0.3 and normalize by 0.3 to obtain PRO AUC. 
\\
d) \textbf{Inference Throughput}: To quantify runtime we feed the entire test split through each model and measure the total wall-clock time. We compute FPS per category and then average over all 15 categories.

\section{Results and Discussion}
\label{sec:results}

\begin{table*}[htbp]
    \caption{Average performance over all 15 MVTec AD categories.}
    \centering
    \label{tab:mvtec-avg}
    \begin{tabular}{lcccccc}
        \toprule
        \textbf{Method} & \textbf{Img AUC} & \textbf{Img AP} & \textbf{Px AUC} & \textbf{Px AP} & \textbf{PRO} & \textbf{FPS} \\
        \midrule
        AE-MSE & 0.708 & 0.859 & 0.733 & 0.152 & 0.417 & 19.8 \\
        D3R-MSE & 0.720 & 0.867 & 0.738 & \textbf{0.177} & 0.441 & \textbf{21.5} \\
        D3R-FFT & 0.706 & \textbf{0.867} & \textbf{0.751} & 0.166 & \textbf{0.468} & 20.3 \\
        D3R-FFT-SSIM & 0.657 & 0.840 & 0.625 & 0.127 & 0.346 & 20.5 \\
        \midrule
        PaDiM (ResNet-18) & 0.894 & 0.943 & 0.964 & 0.443 & 0.733 & 21.7 \\
        STFPM (ResNet-18) & 0.599 & 0.797 & 0.822 & 0.171 & 0.471 & 20.7 \\
        PatchCoreLite (ResNet-18) & \textbf{0.937} & \textbf{0.969} & 0.940 & \textbf{0.476} & \textbf{0.854} & 7.7 \\
        \bottomrule
    \end{tabular}
\end{table*}

\subsection{Average Performance over All Categories}
Several patterns become visible when analyzing the data at this basic level. The system achieves superior results in all evaluation metrics when it switches from AE-MSE to D3R-MSE. The healing task improves both PRO AUC and pixel AP performance because it increases average PRO AUC from 0.417 to 0.441 and pixel AP from 0.152 to 0.177 while maintaining a slight increase in FPS. The model achieves better results in normal-texture modeling because it develops the ability to eliminate synthetic corruptions during training instead of using basic input duplication. Adding the FFT term (D3R-FFT) further improves pixel ROC AUC (from 0.738 to 0.751) and PRO AUC (from 0.441 to 0.468) compared to D3R-MSE. The spectral constraint affects the accuracy of defective image detection localization because it does not impact the image-level evaluation results. The SSIM-augmented configuration (D3R-FFT-SSIM) does not fare as well in this setting. Its average PRO AUC and pixel ROC AUC drop below those of D3R-FFT. Intuitively, combining MSE, FFT and SSIM may overconstrain the reconstructions, making the model less tolerant to harmless variations in lighting or texture and thus more prone to false alarms. Not surprisingly, feature-embedding methods with ResNet-18 backbones, particularly PatchCoreLite, achieve the highest average scores. PatchCoreLite reaches 0.937 image ROC AUC and 0.854 PRO AUC on average, but at a lower throughput of about 7.7 FPS due to the overhead of feature storage and neighborhood search.

\begin{figure}[htbp]
    \centering
    \begin{subfigure}[b]{\columnwidth}
        \centering
        \includegraphics[width=0.75\linewidth]{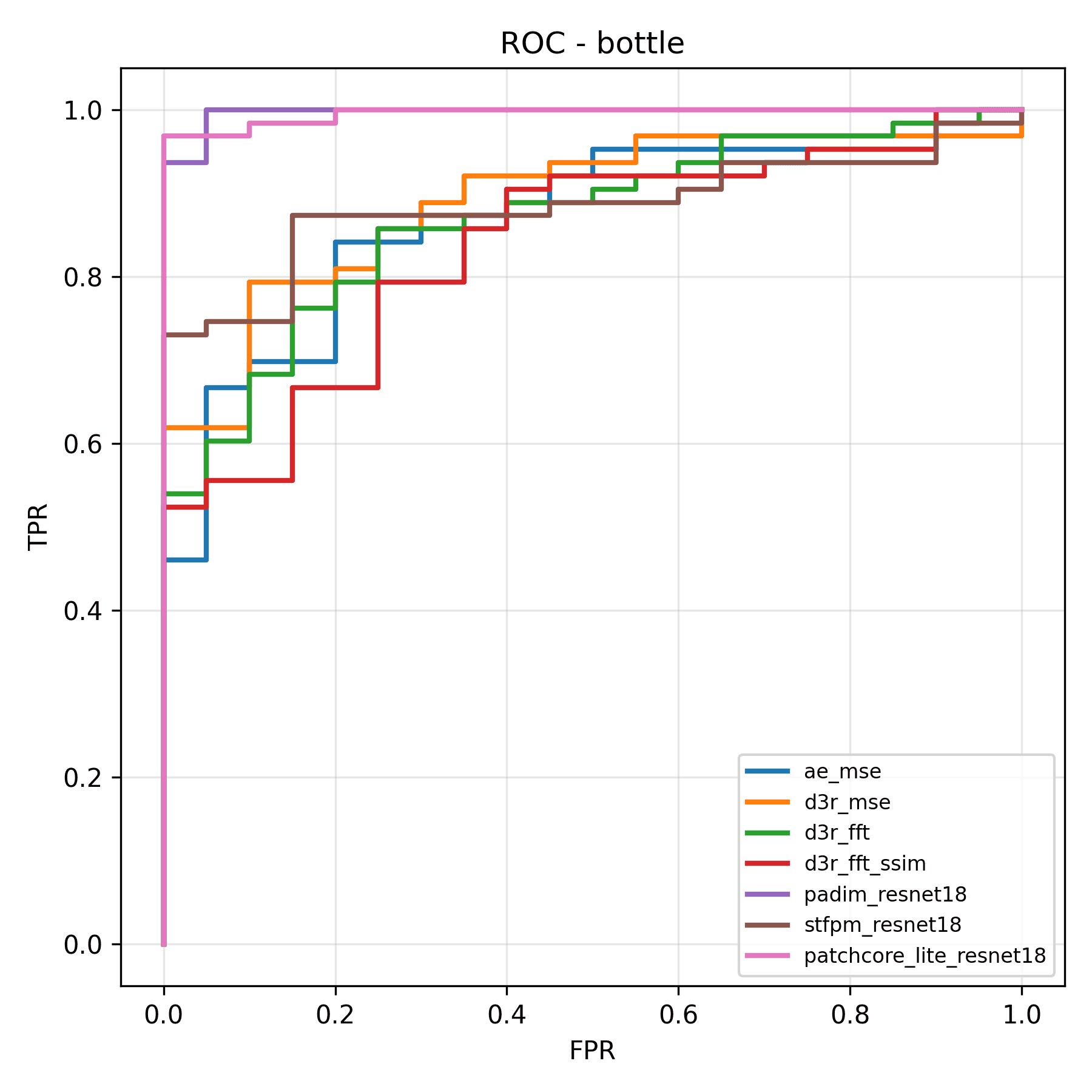}
        \caption{Bottle}
    \end{subfigure}
    \vspace{0.1cm}
    \begin{subfigure}[b]{\columnwidth}
        \centering
        \includegraphics[width=0.75\linewidth]{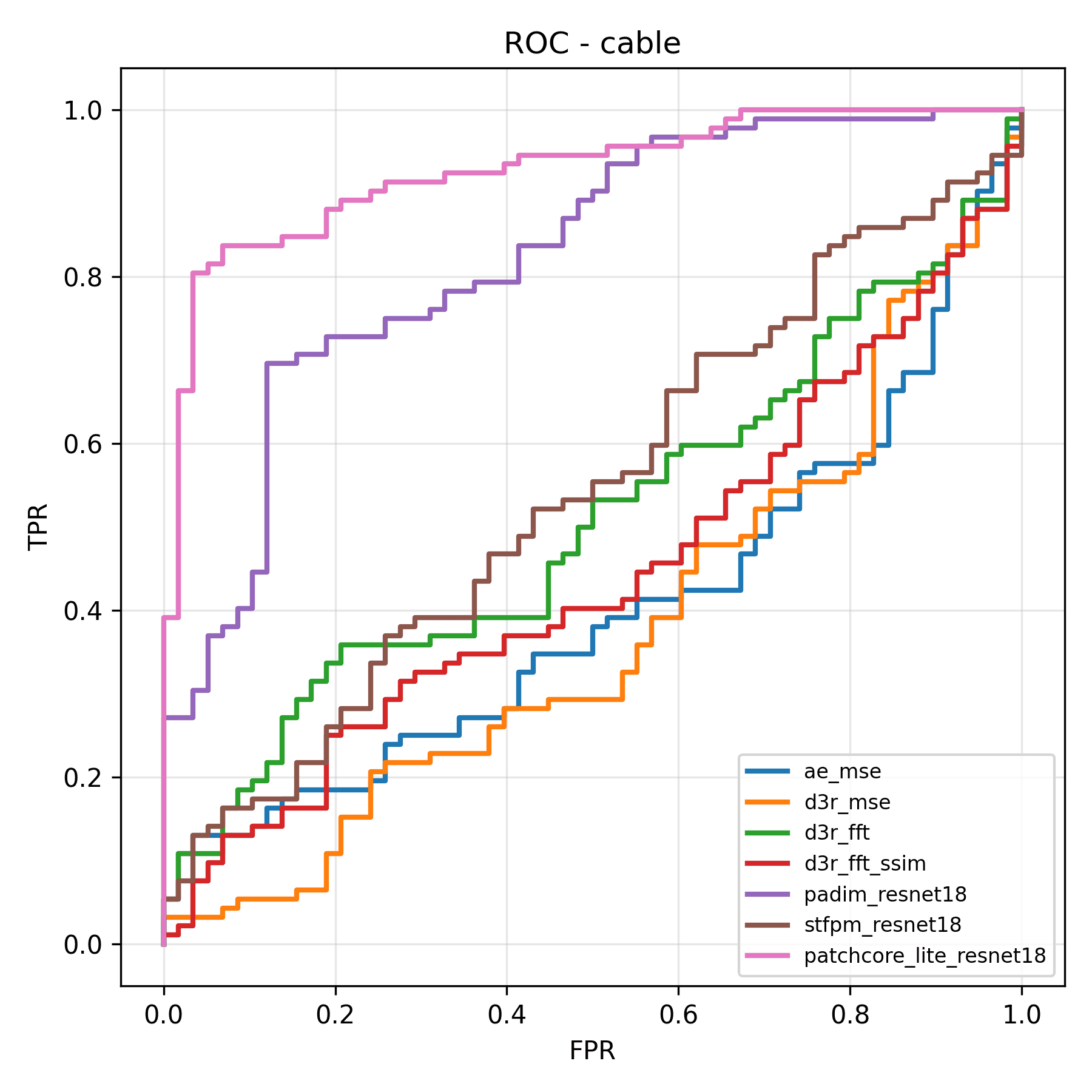}
        \caption{Cable}
    \end{subfigure}
    \vspace{0.1cm}
    \begin{subfigure}[b]{\columnwidth}
        \centering
        \includegraphics[width=0.75\linewidth]{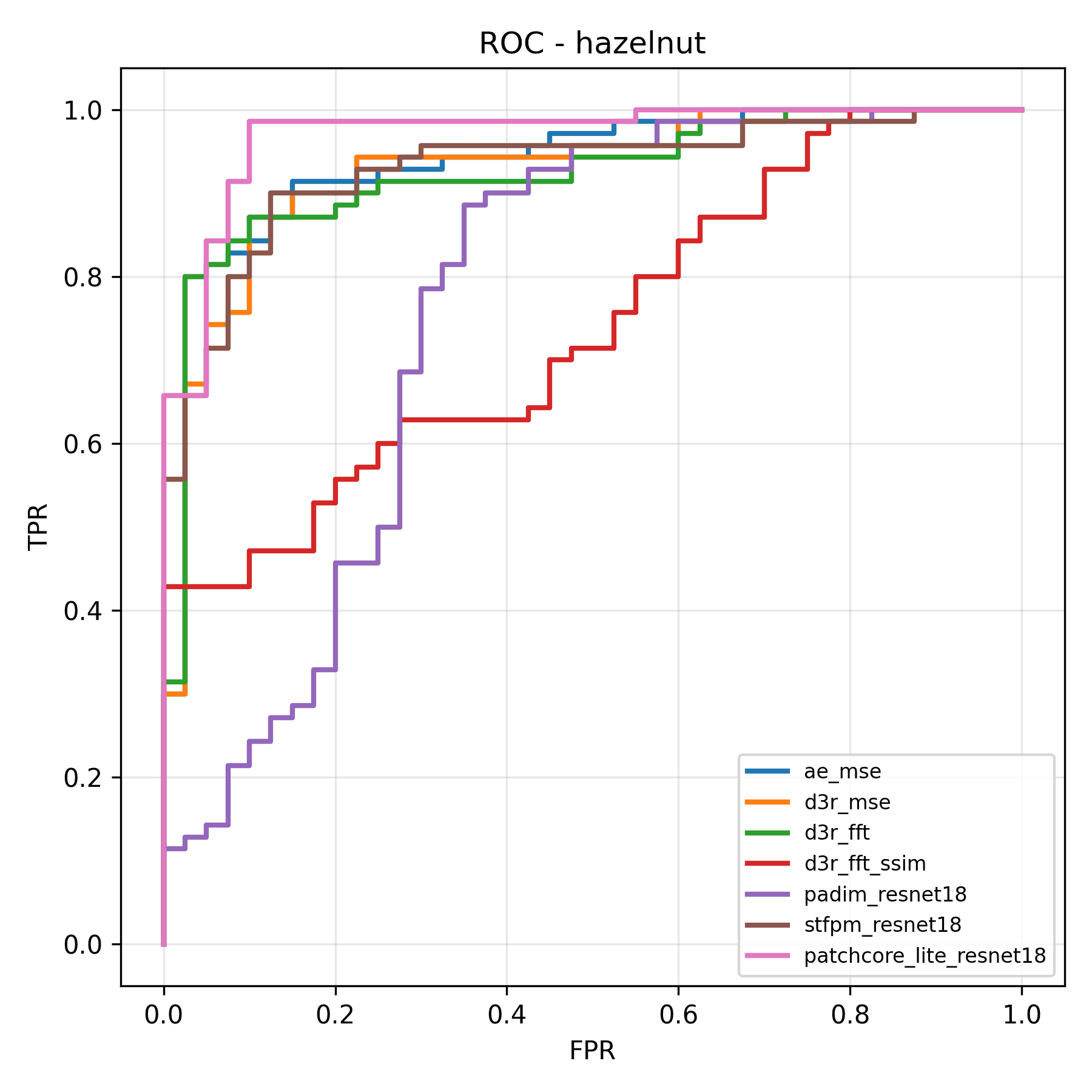}
        \caption{Hazelnut}
    \end{subfigure}
    \caption{Image-level ROC curves on three representative categories (bottle, cable, hazelnut) from the multi-category benchmark. Each plot is taken directly from the benchmark run and compares reconstruction-based methods with feature-embedding baselines.}
    \label{fig:roc_multiclass}
\end{figure}

\subsection{Object vs Texture Categories}
To better understand where D3R-FFT is most helpful, we now look at specific categories that illustrate two different regimes: Object-like categories such as hazelnut, where the model primarily reconstructs a compact foreground object; and Texture-like categories such as leather, where the goal is to reproduce a homogeneous but rich surface pattern over the whole field of view.

\subsubsection{Hazelnut: Reconstructing a Compact Object}
Here the task is to reconstruct individual hazelnuts with various small defects (cracks, holes, contamination) against a relatively simple background. A few remarks are in order: All reconstruction variants achieve strong image-level detection, with image ROC AUC close to or above 0.93. The coarse separation between normal and defective hazelnuts is therefore not the main difficulty. The D3R-MSE configuration slightly improves pixel ROC AUC and Pixel AP over AE-MSE, reflecting the benefit of the healing task even for an object-centric category. D3R-FFT trades some pixel ROC AUC and Pixel AP for a notable increase in PRO AUC (from 0.606 to 0.687). In other words, its anomaly maps cover defect regions more completely, even if the overall ranking of pixels by anomaly score is somewhat less sharp. The improved PRO is valuable in contexts where missing part of a defect is more problematic than raising a few extra false positives. Feature-based methods, and PatchCoreLite in particular, still dominate on Hazelnut. The model achieves high scores at both image and pixel levels because it contains powerful semantic information from the pre-trained ResNet-18 and its capability to detect small differences in feature space.

\begin{table}[htbp]
    \caption{Hazelnut category: image- and pixel-level performance.}
    \centering
    \label{tab:hazelnut}
    \resizebox{\columnwidth}{!}{%
    \begin{tabular}{lccccc}
        \toprule
        \textbf{Method} & \textbf{Img AUC} & \textbf{Img AP} & \textbf{Px AUC} & \textbf{Px AP} & \textbf{PRO} \\
        \midrule
        AE-MSE & 0.936 & 0.961 & 0.914 & 0.468 & 0.603 \\
        D3R-MSE & 0.928 & 0.956 & \textbf{0.925} & \textbf{0.490} & 0.606 \\
        D3R-FFT & 0.923 & 0.956 & 0.882 & 0.428 & 0.687 \\
        D3R-FFT-SSIM & 0.739 & 0.858 & 0.858 & 0.260 & 0.616 \\
        \midrule
        PaDiM & 0.766 & 0.816 & 0.977 & 0.468 & 0.633 \\
        STFPM & 0.930 & 0.963 & 0.966 & 0.437 & 0.813 \\
        PatchCoreLite & \textbf{0.970} & \textbf{0.982} & 0.974 & 0.474 & \textbf{0.911} \\
        \bottomrule
    \end{tabular}%
    }
\end{table}

\begin{figure}[htbp]
    \centering
    \begin{subfigure}[b]{0.45\columnwidth}
        \includegraphics[width=\linewidth]{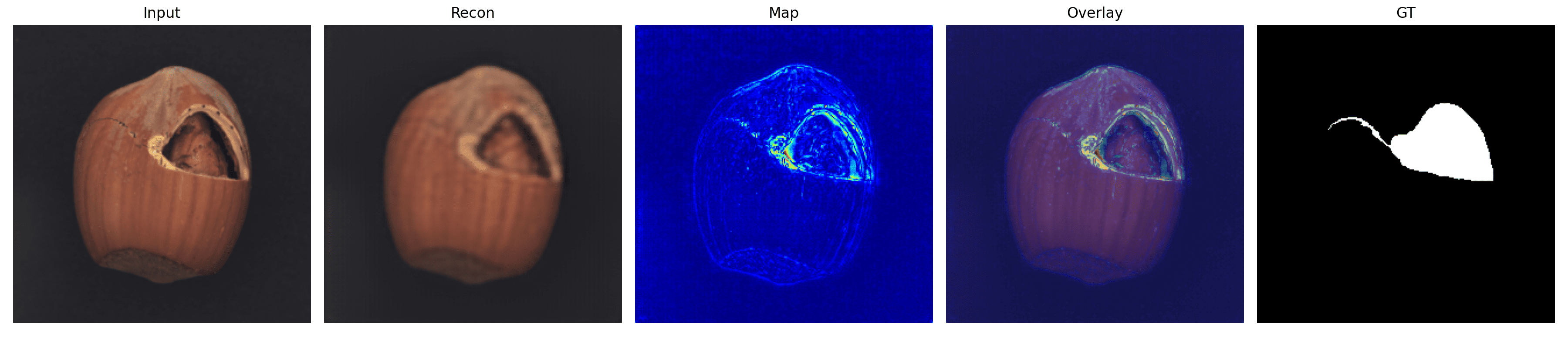}
        \caption{AE-MSE}
    \end{subfigure}
    \hfill
    \begin{subfigure}[b]{0.45\columnwidth}
        \includegraphics[width=\linewidth]{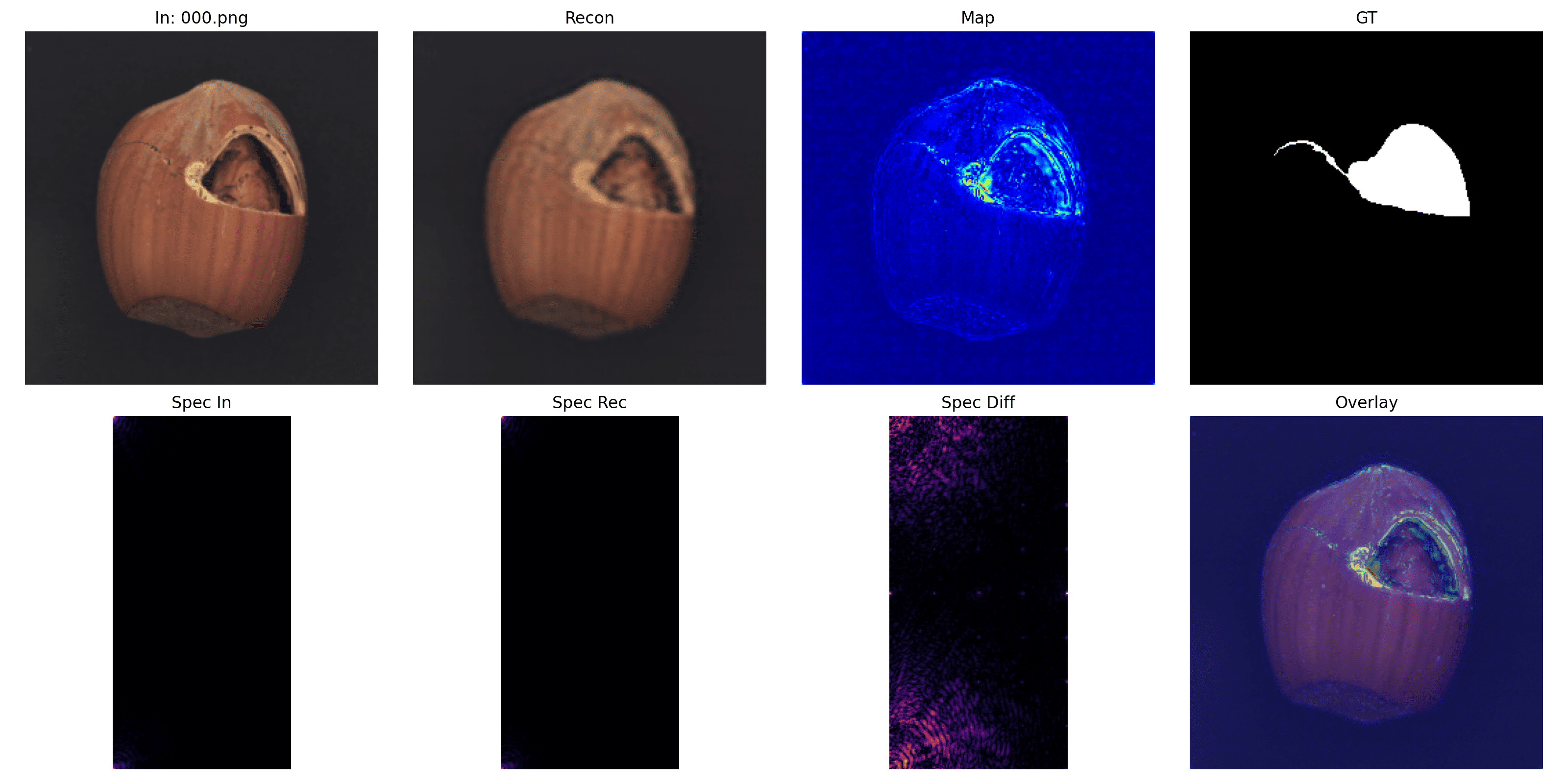}
        \caption{D3R-MSE}
    \end{subfigure}

    \vspace{0.1cm}

    \begin{subfigure}[b]{0.45\columnwidth}
        \includegraphics[width=\linewidth]{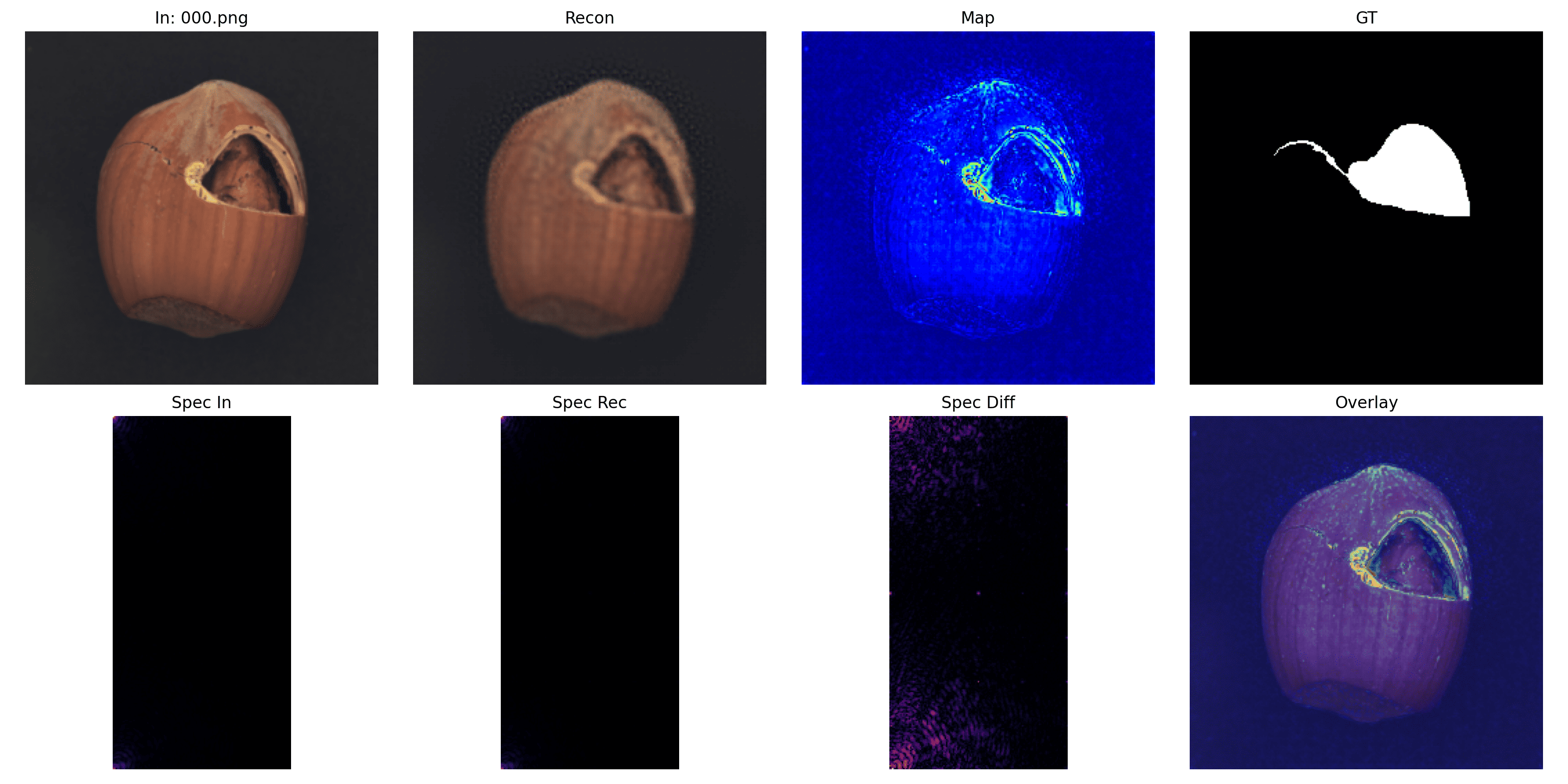}
        \caption{D3R-FFT}
    \end{subfigure}
    \hfill
    \begin{subfigure}[b]{0.45\columnwidth}
        \includegraphics[width=\linewidth]{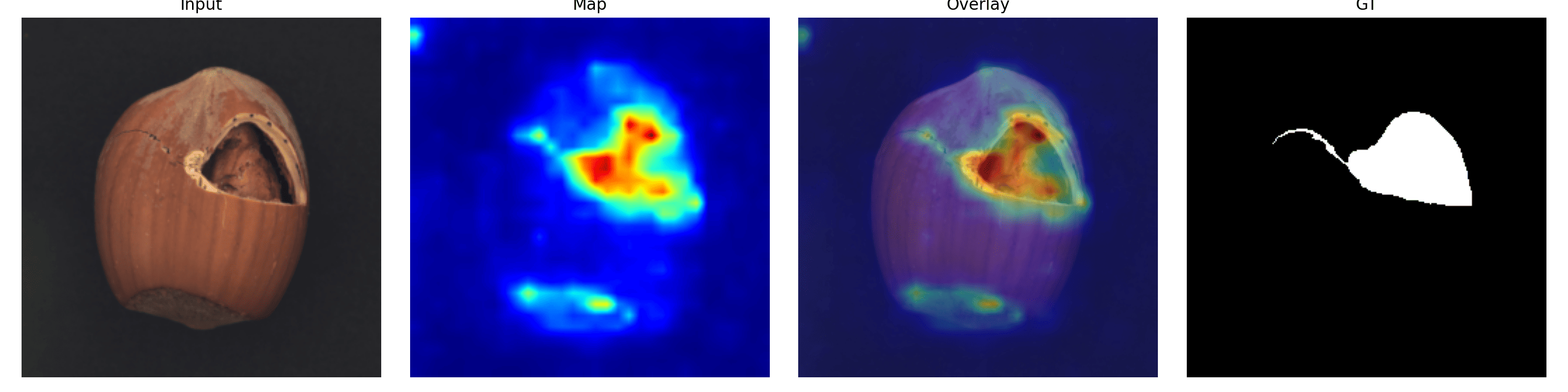}
        \caption{PaDiM}
    \end{subfigure}

    \vspace{0.1cm}

    \begin{subfigure}[b]{0.45\columnwidth}
        \includegraphics[width=\linewidth]{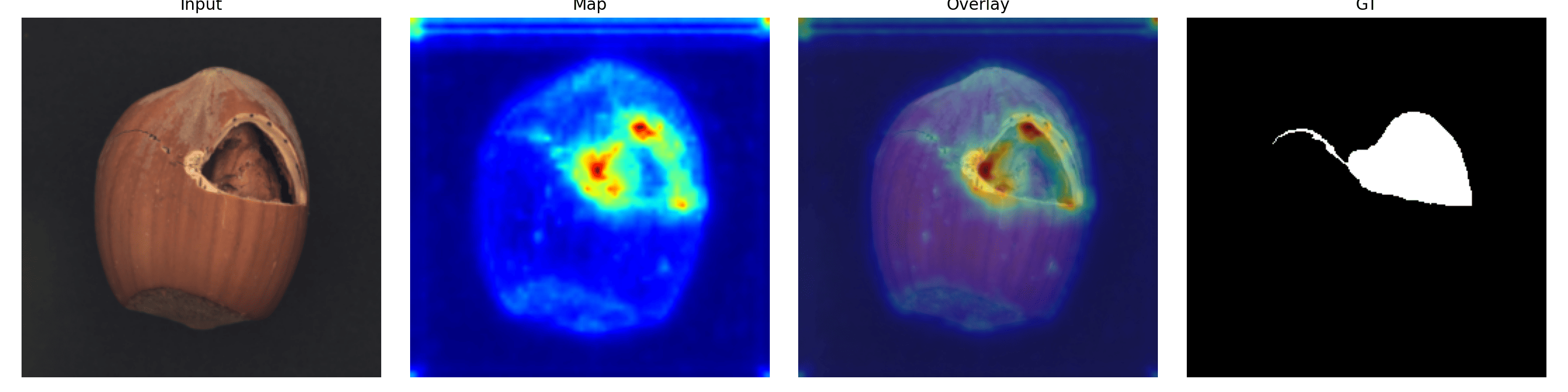}
        \caption{STFPM}
    \end{subfigure}
    \hfill
    \begin{subfigure}[b]{0.45\columnwidth}
        \includegraphics[width=\linewidth]{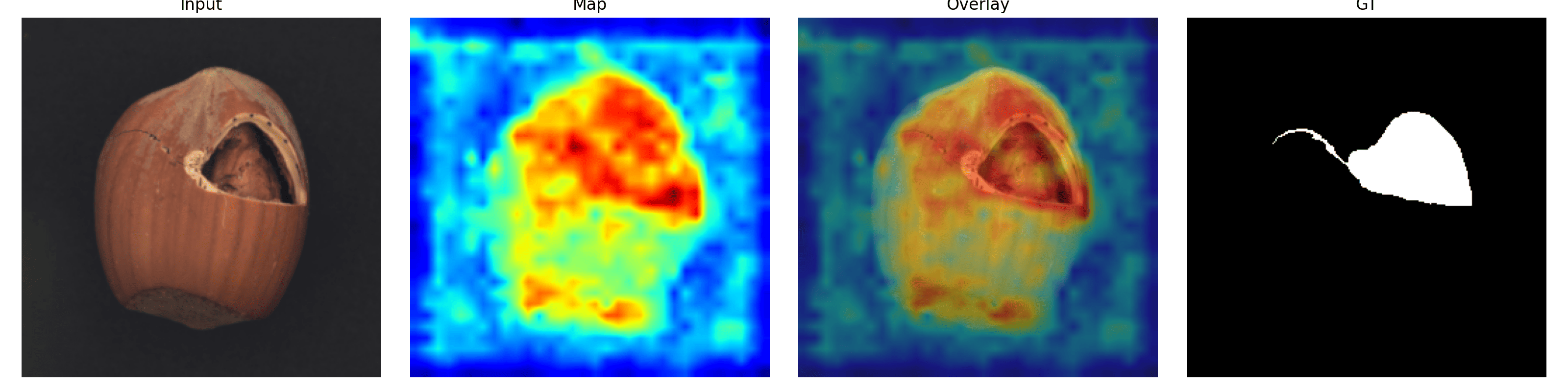}
        \caption{PatchCoreLite}
    \end{subfigure}
    \caption{Qualitative comparison on a Hazelnut test sample. Each panel shows the input, reconstruction (if applicable), anomaly map and overlay.}
    \label{fig:hazelnut_methods_benchmark}
\end{figure}

\subsubsection{Leather: Reconstructing a Surface Texture}
The leather category is of a different nature: there is no single foreground object, only a textured surface where defects manifest as local distortions, discolorations or cuts. Here the behaviour of the dual-domain losses is more subtle. The healing formulation (D3R-MSE) again improves over AE-MSE in all metrics, particularly in PRO AUC and Pixel AP, which is consistent with the idea that the network better captures the distribution of normal texture patches. Unlike Hazelnut, the simple FFT-augmented variant (D3R-FFT) does not clearly dominate here. While PRO AUC marginally increases from 0.467 to 0.471, Pixel AP actually drops. This suggests that enforcing global spectral similarity alone is not sufficient for a class where local texture statistics vary appreciably over the surface. Interestingly, the SSIM-augmented configuration (D3R-FFT-SSIM) performs best among reconstruction-based methods on Leather, with the highest pixel ROC AUC, Pixel AP and PRO AUC. The SSIM term acts exactly on local contrast and structure, which seems particularly well suited for surface-like categories. Feature-embedding methods remain a strong upper bound, with near-perfect image-level detection and high PRO AUC. PatchCoreLite achieves a PRO AUC of 0.880 while still localizing defects fairly sharply.

\begin{table}[htbp]
    \caption{Leather category: image- and pixel-level performance.}
    \centering
    \label{tab:leather}
    \resizebox{\columnwidth}{!}{%
    \begin{tabular}{lccccc}
        \toprule
        \textbf{Method} & \textbf{Img AUC} & \textbf{Img AP} & \textbf{Px AUC} & \textbf{Px AP} & \textbf{PRO} \\
        \midrule
        AE-MSE & 0.846 & 0.945 & 0.748 & 0.113 & 0.445 \\
        D3R-MSE & 0.849 & 0.946 & 0.790 & 0.171 & 0.467 \\
        D3R-FFT & 0.659 & 0.867 & 0.772 & 0.061 & 0.471 \\
        D3R-FFT-SSIM & 0.690 & 0.898 & \textbf{0.885} & \textbf{0.222} & \textbf{0.599} \\
        \midrule
        PaDiM & \textbf{1.000} & \textbf{1.000} & 0.992 & 0.411 & 0.863 \\
        STFPM & 0.660 & 0.881 & 0.965 & 0.286 & 0.736 \\
        PatchCoreLite & \textbf{1.000} & \textbf{1.000} & \textbf{0.996} & 0.539 & \textbf{0.880} \\
        \bottomrule
    \end{tabular}%
    }
\end{table}

\begin{figure}[htbp]
    \centering
    \begin{subfigure}[b]{0.95\columnwidth}
        \includegraphics[width=\linewidth]{d3r_fft_00_000.png}
        \caption{Hazelnut (object-like, D3R-FFT)}
    \end{subfigure}
    \vspace{0.15cm}
    \begin{subfigure}[b]{0.95\columnwidth}
        \includegraphics[width=\linewidth]{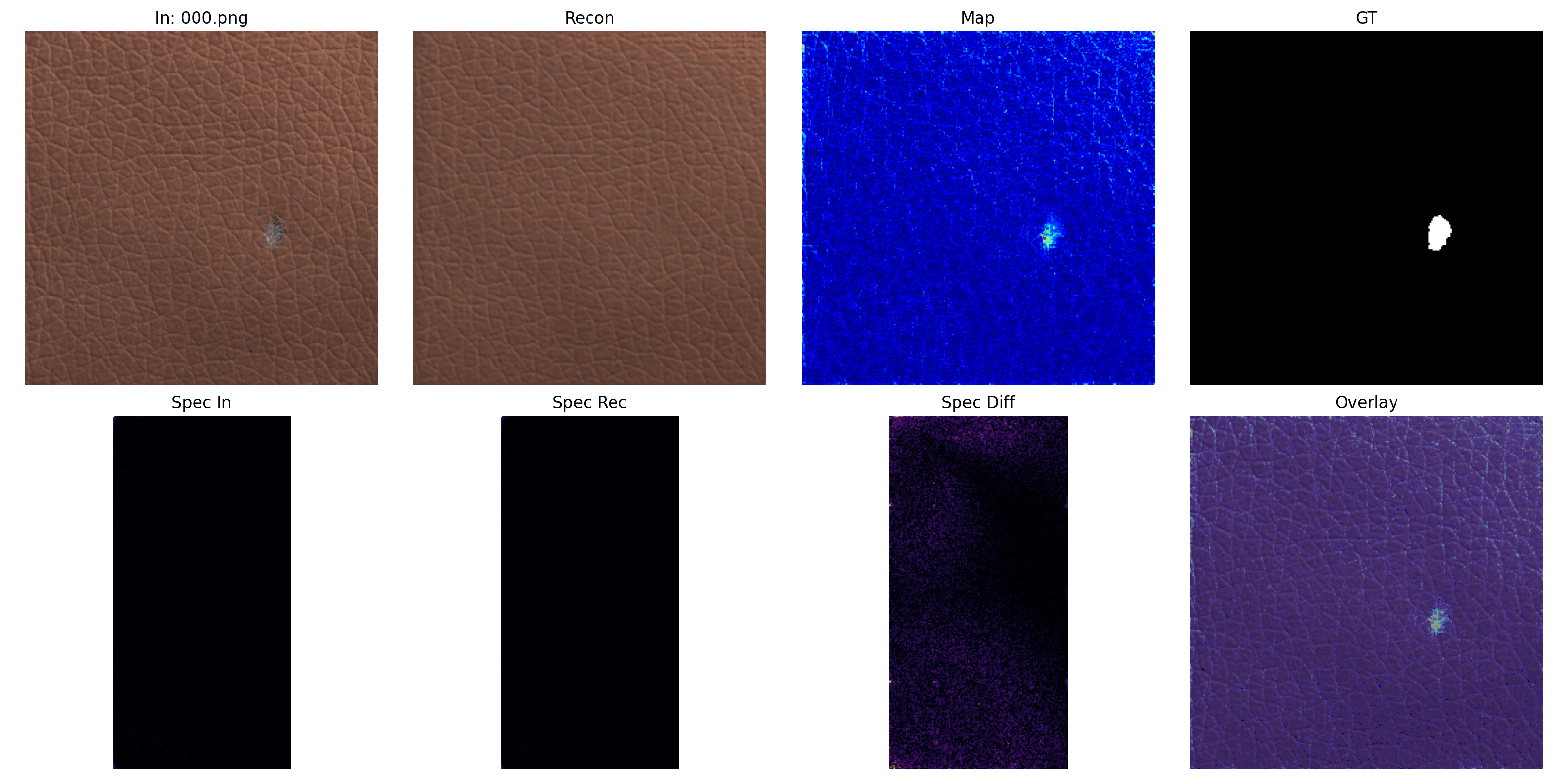}
        \caption{Leather (texture-like, D3R-FFT-SSIM)}
    \end{subfigure}
    \caption{D3R reconstructions and anomaly maps for an object-centric (hazelnut) and a texture-centric (leather) category. The same dual-domain training behaves differently depending on whether the model primarily reconstructs an object or a surface.}
    \label{fig:object_vs_texture_d3r}
\end{figure}

\subsection{Selected Categories: PRO AUC Comparison}
To give a broader view of how methods perform on individual categories, we analyze PRO AUC values for five representative classes, mixing textures and objects. Several observations can be drawn. On Tile, Wood and Pill-categories where anomalies are mainly local disruptions in an otherwise regular texture or object surface the progression from AE-MSE to D3R-MSE to D3R-FFT shows a steady improvement in PRO AUC. The FFT term adds between 0.03 and 0.10 absolute PRO AUC over D3R-MSE. On Screw, which has complex geometry and thin defects along the threads, PRO AUC jumps from 0.356 (AE-MSE) to 0.588 (D3R-FFT). Here the spectral constraint appears to help the autoencoder keep track of the fine thread pattern and thus accentuate deviations. Carpet behaves differently. While D3R-MSE improves over AE-MSE, D3R-FFT slightly reduces PRO AUC. Carpet defects can be large, irregular regions where global spectral matching is less informative than local structure, so the extra frequency penalty may not help. PatchCoreLite dominates across all five categories in terms of PRO AUC, as expected from its use of strong pre-trained features. However, the relative improvements brought by D3R-FFT over AE-MSE and D3R-MSE are substantial in several cases, narrowing the gap with PatchCoreLite while using a markedly simpler backbone.

\begin{table}[htbp]
    \caption{PRO AUC on selected MVTec AD categories (higher is better).}
    \centering
    \label{tab:pro-multi}
    \begin{tabular}{lcccc}
        \toprule
        \textbf{Category} & \textbf{AE-MSE} & \textbf{D3R-MSE} & \textbf{D3R-FFT} & \textbf{PatchCoreLite} \\
        \midrule
        Tile      & 0.412 & 0.449 & 0.453 & 0.739 \\
        Wood      & 0.495 & 0.555 & 0.608 & 0.895 \\
        Pill      & 0.534 & 0.619 & 0.721 & 0.848 \\
        Carpet    & 0.255 & 0.337 & 0.316 & 0.923 \\
        Screw     & 0.356 & 0.427 & 0.588 & 0.890 \\
        \bottomrule
    \end{tabular}
\end{table}

\subsection{Summary of Empirical Findings}
Putting all results together, three main points emerge:

The self-supervised healing task consistently improves reconstruction-based anomaly detection over a plain autoencoder across almost all categories. This effect is visible in both average metrics and per-category analyses.

The FFT magnitude loss is most beneficial on categories where anomalies primarily disturb high-frequency content against a relatively structured background. It significantly improves PRO AUC on Hazelnut, Tile, Pill, Wood and Screw, at essentially no cost in inference speed.

No single configuration of dual-domain losses is uniformly optimal across all 15 categories. While D3R-FFT is a strong and simple default, the SSIM-augmented variant can outperform it on certain texture-dominated classes such as Leather. At the same time, feature-embedding methods remain the reference choice where pre-trained backbones and higher compute budgets are acceptable.

From a practical standpoint, D3R-Net provides a family of reconstruction-based detectors that are easy to train, require only normal data, and offer a tunable balance between spatial fidelity and spectral consistency. The experiments across the full MVTec AD dataset indicate that, with carefully chosen dual-domain regularization, such models can serve as a competitive alternative in scenarios where heavy pre-trained feature extractors are not desirable or feasible.

\section{Conclusion}
\label{sec:concl}

We presented D3R-Net, a dual-domain denoising reconstruction approach for industrial anomaly detection. By framing training as a healing task on synthetically corrupted normal images and adding an FFT-based magnitude loss on top of a simple convolutional autoencoder, the method strengthens the sensitivity of reconstruction-based models to subtle structural defects. On the MVTec AD Hazelnut benchmark, the FFT-regularized variant substantially improves localization consistency (PRO AUC) over an MSE-only autoencoder while maintaining strong image-level ROC AUC. Embedded into a unified evaluation pipeline across all MVTec categories, D3R-Net achieves consistent gains in pixel ROC AUC and PRO AUC over basic reconstruction baselines at comparable inference speed. The system provides its users with a special benefit which helps them when their system has limited available resources. The method differs from current state-of-the-art feature embedding approaches because it trains a small model from beginning to end without using large backbones or external training data. The model functions as a basic framework which helps users understand edge computing operations and situations that require different pre-training methods than ImageNet. Future work may explore combining dual-domain losses with more expressive decoders or integrating them directly into feature embedding methods. Furthermore, while global FFT losses are effective for periodic structures, localized frequency analysis techniques, such as Short-Time Fourier Transforms (STFT) or wavelet-based losses, could be investigated to better handle non-stationary texture anomalies and further bridge the gap to larger models.


\end{document}